\documentclass[conference,onecolumn,10pt]{IEEEtran}
\usepackage{amsmath,amsfonts}
\usepackage{color}
\usepackage{array}
\usepackage[colorlinks,urlcolor=blue,linkcolor=blue,citecolor=blue]{hyperref}
\usepackage{textcomp}
\usepackage{stfloats}
\usepackage{url}
\usepackage{times}
\usepackage{verbatim}
\usepackage{graphicx}
\usepackage{cite}
\usepackage{longtable}
\usepackage[utf8]{inputenc}
\usepackage[T1]{fontenc}
\usepackage{lipsum}
\usepackage[left=.6in,right=.6in,top=1.2 in,bottom=1.2 in]{geometry}
\usepackage{comment}
\usepackage{booktabs} 

\usepackage{algorithm}
\usepackage{algpseudocode}
\usepackage{graphicx}
\usepackage{subcaption}
\usepackage{amsmath}

\begin{document}
\makeatletter
\newcommand{\nosemic}{\renewcommand{\@endalgocfline}{\relax}}
\newcommand{\dosemic}{\renewcommand{\@endalgocfline}{\algocf@endline}}
\newcommand{\pushline}{\Indp}
\newcommand{\popline}{\Indm\dosemic}
\let\oldnl\nl
\newcommand{\nonl}{\renewcommand{\nl}{\let\nl\oldnl}}
\makeatother
\title{A Full Compression Pipeline for Green Federated Learning in Communication-Constrained Environments}

\author{\IEEEauthorblockN{Elouan Colybes\IEEEauthorrefmark{1}, Shirin Salehi\IEEEauthorrefmark{1}, and Anke Schmeink\IEEEauthorrefmark{1} }

\IEEEauthorblockA{\IEEEauthorrefmark{1} Chair of Information
Theory and Data Analytics (INDA), RWTH Aachen University, Aachen, Germany}    

    elouan.colybes@rwth-aachen.de, \{shirin.salehi, anke.schmeink\}@inda.rwth-aachen.de\vspace{-3mm}
}


\maketitle

\begin{abstract}
Federated Learning (FL) enables collaborative model training across distributed clients without sharing raw data, thereby preserving privacy. However, FL often suffers from significant communication and computational overhead, limiting its scalability and sustainability. In this work, we introduce a Full Compression Pipeline (FCP) for FL in communication-constrained environments. FCP integrates three complementary deep compression techniques—pruning, quantization, and Huffman encoding—into a unified end-to-end framework. By compressing local models and communication payloads, FCP substantially reduces transmission costs and resource consumption while maintaining competitive accuracy. To quantify its impact, we develop an evaluation framework that captures both communication and computation overheads as a unified model cost, allowing a holistic assessment of efficiency trade-offs.
The pipeline is evaluated in an independent and identically distributed (IID) and non-IID data setting. In one representative scenario, training a ResNet-12 model on the CIFAR-10 dataset with ten clients and a 2~Mbps bandwidth, the FCP achieves more than 11$\times$ reduction in model size, with only a 2\% drop in accuracy compared to the uncompressed baseline. This results in an FL training that is more than 60\% faster. 
\end{abstract}

\section{Introduction}
Recent advances in artificial intelligence (AI), particularly large language models (LLM), have transformed the perception and applications of AI across domains. This progress, enabled by increasing computational power, has led to unprecedented model complexity but also to significant energy demands and environmental concerns. The emerging paradigm of green AI advocates for efficient, sustainable learning systems~\cite{schwartz_green_2020, 10251541}, contrasting with red AI, which prioritizes performance regardless of computational cost.

Within machine learning (ML), federated learning (FL)~\cite{mcmahan2023communicationefficientlearningdeepnetworks} offers a distributed training paradigm where clients collaboratively train a global model without sharing raw data, enhancing privacy. However, despite these benefits, FL still suffers from significant communication and computational overhead due to frequent model updates. To address these challenges, we propose integrating deep compression techniques~\cite{han2016deepcompressioncompressingdeep}—pruning, quantization, and Huffman encoding—into an end-to-end \textit{Full Compression Pipeline} (FCP) for FL. FCP aims to improve communication efficiency, scalability, and sustainability while maintaining accuracy, aligning with green AI principles.

The main goals of this paper are: (1) develop a flexible FL simulation framework supporting compression techniques; (2) provide an analytical characterization of the proposed FCP, including compression ratio, compression loss, and complexity analysis; (3) design a comprehensive FL evaluation framework to gauge computation and communication overheads as a unified model cost; and (4) demonstrate FCP’s generalizability across datasets and data distributions. The remainder of this paper reviews related work (Sec.~II), describes the system setup (Sec.~III), details the FCP (Sec.~IV), presents results (Sec.~V), and concludes (Sec.~VI).

\section{Related Works}

Communication efficiency remains a major bottleneck in FL due to frequent model exchanges over constrained networks. Shahid et al.~\cite{shahid2021communicationefficiencyfederatedlearning} highlight that bandwidth limits, device heterogeneity, and non-IID data hinder scalability, and survey mitigation strategies such as client selection, local updates, compression, and decentralization. Grativol et al.~\cite{ribeiro2023federatedlearningcompressiondesigned} integrate pruning and quantization-aware training (QAT) into the FL pipeline, but only as separate components to achieve a reduction in message size with small accuracy loss.
Similarly, FedZip~\cite{malekijoo2021fedzipcompressionframeworkcommunicationefficient} combines Top-$z$ sparsification, $k$-means quantization, and encoding to reduce the communication costs.  
ResFed~\cite{song2023resfed} transmits compressed residuals predicted by local and server models, to yield communication savings while maintaining accuracy. CMFL~\cite{wang2019overhead} filters non-aligned client updates to improve efficiency, and FedCompress~\cite{tsouvalas2024communication} employs adaptive clustering and server-side distillation for reduction. 

While existing approaches promote lightweight communication, they typically lack analytical evaluation of compression techniques in terms of compression ratio, accuracy loss, and computational complexity. Moreover, they often overlook the additional overhead introduced by these methods. In contrast, we introduce a unified cost model that analytically captures their joint impact on FL performance.

\begin{figure*}[t]
    \centering
    \includegraphics[width=1\linewidth]{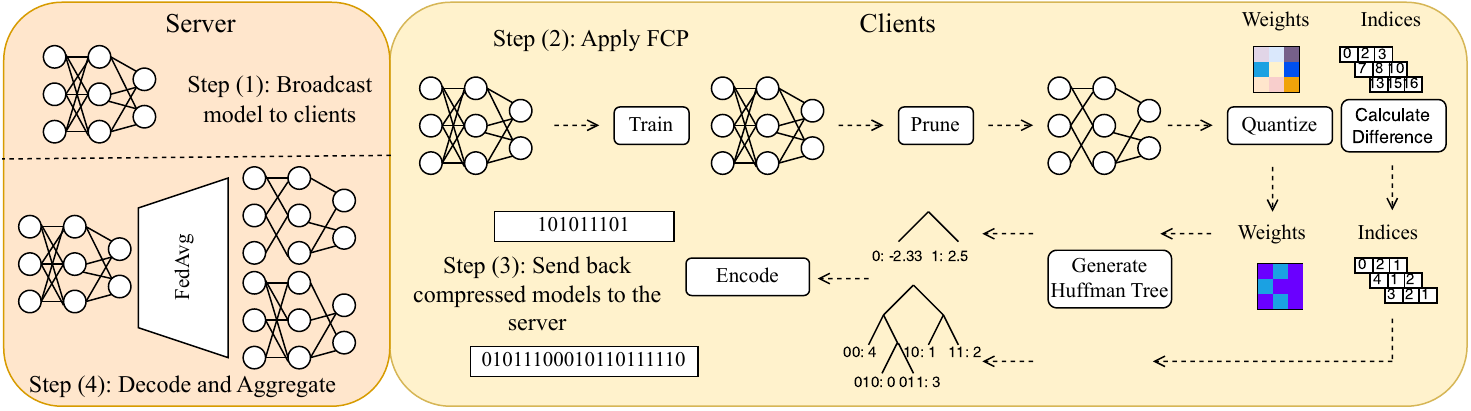}
    \caption{Our full compression pipeline to reduce communication overhead in federated learning }
    \label{fig:FCP}
\end{figure*}

\section{System setup}

We consider a server tasked with training a global model using the FL paradigm. The server coordinates a set of clients $C$, each holding private local data. At each communication round $i$, the server selects a subset $S \subset C$, sends them the current global model, and receives their locally updated parameters $w_s^{(i)}$. The server then aggregates these updates using the Federated Averaging (FedAvg) algorithm~\cite{mcmahan2017communication} to refine the global model through weighted averaging.
Clients, simulating edge devices, perform local training and evaluation before transmitting their compressed updates to the central server. 
In our setup, we apply deep compression techniques after local training and before model upload, while maintaining full control over optimization hyperparameters such as learning rate, batch size, and momentum. 
In FL, compression can be applied upstream (client-to-server), downstream (server-to-client), or bidirectionally~\cite{shah2023mcfcefl}, though most approaches focus on the first two~\cite{malekijoo2021fedzipcompressionframeworkcommunicationefficient}. Upstream compression reduces server-side bottlenecks by minimizing update transmission, whereas downstream compression adds decoding overhead and may propagate errors to clients. Hence, we apply compression only upstream to balance efficiency and model integrity.

\section{Proposed full compression pipeline for FL}
The goal of the proposed FCP is to minimize communication costs in FL while maintaining model fidelity. To achieve this, we adapt established compression techniques—pruning, quantization, and Huffman encoding, in the same order as introduced by Han et al. \cite{han2016deepcompressioncompressingdeep}—to the federated setting by integrating them sequentially within a single, layer-wise pass. Executing them in this order present multiple advantages, as detailed later. Performing all compression steps per layer in one loop improves computational efficiency compared to applying each method independently across all layers.
We then derive explicit expressions for the communication compression ratio, analytical bounds on accuracy degradation due to compression noise, and a per-layer complexity characterization for client-side operations.
The FCP pipeline is illustrated in Fig.~\ref{fig:FCP} and Algorithm \ref{alg:fl-fcp}.

\algrenewcommand{\algorithmicrequire}{\textbf{Input:}}
\algrenewcommand{\algorithmicensure}{\textbf{Output:}}

\subsection{Compression Pipeline}

\subsubsection{Pruning}
We adopt unstructured, model-wise pruning, which removes individual weights with small magnitudes across the entire model. Our ablation studies indicate that this approach achieves a favorable trade-off between model sparsity and accuracy retention compared to structured or layer-wise pruning. Specifically, we set a global pruning threshold $\lambda$ as the $\gamma$-quantile of all weight magnitudes, so that the smallest $\gamma$ fraction of weights—those contributing least to the model output—are removed. This reduces memory usage and computational cost without significantly affecting model performance. Formally, let $\mathcal{W} = \{w_1, w_2, \ldots, w_N\}$ denote the set of all model weights, and let $\mathcal{W}_{\mathrm{np}} = \{w_i \in \mathcal{W} \mid |w_i| \geq \lambda\}$ be the set of non-pruned weights, so that $|\mathcal{W}_{\mathrm{np}}| = (1-\gamma)N$.

\subsubsection{Post-training quantization (PTQ)}
We adopt codebook quantization (CQ)~\cite{han2016deepcompressioncompressingdeep}, a $k$-means-based method that clusters the reduced set of non-pruned weights $\mathcal{W}_{\text{np}} = \{w_1, \dots, w_{|\mathcal{W}_{\text{np}}|}\}$ into $k$ clusters $\mathcal{C} = \{c_1, \dots, c_k\}$, with each cluster sharing a common value. Weights are not shared across layers. The clustering minimizes the within-cluster sum of squares:
$\underset{\mathcal{C}}{\operatorname{argmin}} \sum_{i=1}^k \sum_{w \in c_i} |w-c_i|^2$.
The number of clusters is typically set by a bit parameter $q$ so that $k = 2^q$. The set of non-pruned and quantized weights is then $\mathcal{W}_{\text{npq}}$, having the same size as $\mathcal{W}_{\text{np}}$, and $\forall~ w \in \mathcal{W}_\text{npq}, \exists~i \in \{1,k\}, ~w = c_i$.
Based on our ablation studies, we adopt layer-wise quantization for better results. With Huffman encoding, fixed-length bit representation is unnecessary, allowing flexible cluster specification.

\subsubsection{Huffman Encoding for Pruned and Quantized Networks}\label{sec:4-huffman}
We apply Huffman encoding to compress pruned and quantized weights in a lossless manner. Each weight value $w~\in \mathcal{W}_\text{npq}$ is replaced by a variable-length code $c(w)$ such that the length $\ell(w) \propto -\log_2 p(w)$, where $p(w)$ is the empirical frequency of $w$ in the layer. This ensures shorter codes for common values and longer codes for rare ones, minimizing storage and memory overhead without affecting accuracy. The encoded data is transmitted to the server.
Following Han et al.~\cite{han2016deepcompressioncompressingdeep}, we Huffman-encode the index differences $\Delta(w)$ corresponding to the positions of nonzero weights to exploit repeated sparsity patterns within each layer. This step further reduces communication cost while preserving exact model reconstruction. The sequence of the three compression techniques is theoretically fixed: pruning prepares the model for subsequent compression steps, while quantization further structures the weight distribution in a way that facilitates more effective entropy encoding.

\begin{algorithm}[tb]
    \caption{FL with Full Compression Pipeline (FCP)}
    \begin{algorithmic}[1]
    \Require $\gamma$: pruning rate  
    \Require $k$: number of clusters for quantization  
    \Require $h$: hyperparameters for local training  
    \Require $R$: number of communication rounds  
    \Require $C$: set of all clients  
    \Require $\mathcal{W}^{(0)}$: initial global model weights  

    \For{each round $i = 1$ to $R$}
        \State randomly select a subset $S \subset C$ 
        \State broadcast $\mathcal{W}^{(i-1)}$ to all $s \in S$ 
        \For{each client $s \in S$ \textbf{in parallel}}
            \State perform local training using $h$ to obtain $\mathcal{W}_s^{(i-1)}$
            \State $\lambda \gets \gamma\text{-quantile}(\mathcal{W}_s^{(i-1)})$     //threshold for the unstructured pruning
                 \hfill
            \For{$\ell = 1,\dots,L$}
                \State 
        Find $\mathcal{W}_{\text{np}} = \{ w_i \in \mathcal{W}_{s,\ell}^{(i-1)} \mid |w_i| \geq \lambda \}$      //set of non-pruned weights
                \State Quantize $\mathcal{W}_{\text{np}}$ in $k$ clusters: $\mathcal{W}_{\text{npq}}$ 
                \State 
                Calculate differences on $\mathcal{W}_{\text{npq}}$: $\{\Delta_i\}_{i=1}^{|\mathcal{W}_{\text{np}}|}$
                \State Huffman encode $\mathcal{W}_{\text{npq}}$, $\{\Delta_i\}_{i=1}^{|\mathcal{W}_{\text{np}}|}$ and send the encrypted values and codebooks
            \EndFor
        \EndFor
        \State server decodes all received updates
        \State compute $n = \sum_{s \in S} n_s$
        \State aggregate:
        \[
            w^{(i)}_{\text{global}} \;=\; \sum_{s \in S} \frac{n_s}{n}\,w_s^{(i)}
        \]
    \EndFor
    \end{algorithmic}
    \label{alg:fl-fcp}
\end{algorithm}

\subsubsection{Huffman decoding}\label{sec:4-huffman-decoding}
At the server, the neural network is reconstructed by reversing client-side encoding: Huffman-encoded data, indices, and codebooks are read, and using the Huffman trees, data and index differences are decoded into 1-D arrays. Index differences are converted to absolute indices via cumulative sums, enabling reconstruction of the sparse matrix (CSR/CSC) and full weight matrices by filling zeros. Once all client models are rebuilt, aggregation methods like FedAvg are applied as usual.

\subsection{Analytical Evaluation}
\subsubsection{Compression Ratio}
During each communication round, clients transmit pruned, quantized, and Huffman encoded model updates to the central server for global aggregation. The communication overhead is measured by the total number of bits transmitted per round, denoted as $\mathcal{B}(\gamma, k)$.
Let the uncompressed model contain $N$ trainable parameters, each represented with $b$ bits. The corresponding baseline transmission cost is $\mathcal{B}_0 = N \cdot b$.
After applying pruning with rate $\gamma$ and quantization with $k$ clusters, and before Huffman encoding, the effective transmission cost becomes 
$\mathcal{B}_{pq}(\gamma, k) = (1 - \gamma)N \cdot \log_2(k)$. 
The final transmission cost for a single client after applying Huffman encoding is given by
\begin{equation}
\mathcal{B}_{pqh}(\gamma, k) = 
\sum_{w \in \mathcal{W}_{\mathrm{npq}}}
\big(\ell(w) + \ell(\Delta(w))\big) \\
\quad + \big(kb + \sum_{j=1}^{k} \ell(c_j)\big) 
+ \big(Mb_{\Delta} + \sum_{r=1}^{M} \ell(\Delta_r)\big).
\end{equation}
The total communication per round is obtained by summing over all participating clients.
 The first term represents the encoded payload (quantized weight values and their index differences).  
The second term corresponds to the codebook for centroids $\text{Codebook}_{\text{centroids}} = \{ (c_j, \ell(c_j)) \mid j = 1, \ldots, k \},$ where $k$ is the number of clusters and $b$ is the number of bits used to represent each centroid.  
The third term accounts for the codebook of distinct index differences, $\text{Codebook}_{\text{idx}} = \{ (\Delta_r, \ell(\Delta_r)) \mid r = 1, \ldots, M \},$ where 
$\{\Delta_r\}_{r=1}^{M}$ is the set of distinct index differences, 
$M = |\{\Delta_r\}|$, and 
$b_{\Delta}$ is the number of bits required to represent each $\Delta_r$.
We can rewrite the encoded payload of quantized weights as \\$\mathcal{B_{\text{weights}}} = (1 - \gamma) N \sum_{j=1}^{k} p(c_j) \, \ell(c_j)$ and their index differences as $\mathcal{B}_{\text{indices}} = (1 - \gamma) N \sum_{r=1}^M p(\Delta_r) \, \ell(\Delta_r)$. Then the corresponding compression ratio $ \frac{\mathcal{B}_{pqh}(\gamma, k)}{\mathcal{B}_0}$ is
\begin{equation}
    H_{comm}(\gamma, k) = \frac{1 - \gamma }{b} \bigg[\sum_{j=1}^{k} p(c_j)\ell(c_j) 
    + \sum_{r=1}^M p(\Delta_r) \ell(\Delta_r)\bigg]\\
    +\frac{k}{N}+\frac{1}{Nb}\sum_{j=1}^{k} \ell(c_j)
    + \frac{Mb_{\Delta}}{Nb}+\frac{1}{Nb}\sum_{r=1}^{M} \ell(\Delta_r).
    \label{comp ratio}
\end{equation}

As $N$ is usually very large, and the codebooks have a relatively small size, the first term is widely dominant. This indicates that the codebook overhead becomes negligible for large models. 

\subsubsection{Compression Loss}

\textbf{Assumptions.} 
We consider an $L$-smooth loss function, assume local stochastic gradients have bounded variance $\sigma_g^2$, 
and model compression (pruning and $k$-means quantization) as an additive error $\epsilon_s$ with bounded energy:
$\mathbb{E}\|\epsilon_s\|^2 \le \sigma_\epsilon^2.$

Let $w_s^{(i)}$ denote the local model of client $s$ at round $i$, and let $\mathcal{Q}(\cdot)$ represent the compression operator. The compressed update can be written as
$\tilde{w}_s^{(i)} = \mathcal{Q}(w_s^{(i)}) = w_s^{(i)} + \epsilon_s^{(i)}$,
where $\epsilon_s^{(i)}$ captures the compression-induced error. Since $k$-means quantization and pruning are generally biased, 
$\mathbb{E}[\epsilon_s^{(i)}] \neq 0,$
and this bias propagates through FedAvg aggregation:
\begin{equation}
w_{\text{global}}^{(i+1)} = \sum_{s \in S} \frac{n_s}{n} \tilde{w}_s^{(i)} 
= \sum_{s \in S} \frac{n_s}{n} w_s^{(i)} + \sum_{s \in S} \frac{n_s}{n} \epsilon_s^{(i)}.
\end{equation}

Under moderate pruning ratios and sufficiently fine quantization (large $k$), the perturbation remains bounded:
$\mathbb{E}\big[\|w_{\text{global}}^{(i+1)} - w_{\text{global}}^{(i+1,\mathrm{uncompressed})}\|^2\big]
\le \sigma_\epsilon^2.$
As long as $\sigma_\epsilon^2$ is small relative to the stochastic gradient variance $\sigma_g^2$, the global model converges with only minor accuracy degradation. 
This additive-noise perspective provides a tractable way to relate compression strength to expected accuracy loss without exhaustive per-layer simulations.

\subsubsection{Computational Complexity}

We compare the computational complexity of FCP with standard FL to quantify the additional cost introduced by pruning, quantization, and encoding.
Let $n_\ell$ denote the number of trainable parameters in layer $\ell$, and $N = \sum_{\ell=1}^{L} n_\ell$ the total model size.

\textbf{Baseline FL.}
For each client, forward and backward propagation as well as parameter updates scale linearly with $N$, giving a per-client, per-round complexity of $O(N)$.
Across $R$ communication rounds with $|S|$ participating clients per round, the total cost is
$O_{\text{total}}^{\text{FL}} = O(R |S| N).$

\textbf{FCP.}
Beyond local training, each client performs model pruning, $k$-means quantization, and Huffman encoding. 
Pruning requires finding the threshold using a selection algorithm, $O(N)$, and scanning all parameters, $O(N)$. 
Quantization on the remaining $(1-\gamma)N$ weights adds $O((1-\gamma)N k I)$, where $k$ is the number of clusters and $I$ is the number of $k$-means iterations. 
Huffman encoding of the quantized weights and index differences contributes $O((1-\gamma)N + k \log k)$ for the values and $O((1-\gamma)N + M \log M)$ for the indices, where $M \ll N$.
Summing over all layers yields a per-client complexity of
$O_{\text{comp}}^{\text{FCP}} = O\big(N + (1-\gamma) N k I + 2(1-\gamma)N + k \log k + M \log M\big).$
Treating $k$ and $I$ as constants, the overall complexity remains linear in $N$, assuming $M \ll N$. 
Thus, the total computational cost over all clients and rounds becomes $O_{\text{total}}^{\text{FCP}} = O(R |S| N).$

\textbf{Discussion.}
Although FCP introduces additional local computations, the overall complexity remains linear in $N$, similar to standard FL.
Since $k$ and $I$ are small constants and $1-\gamma \leq 1$, the extra compression overhead is modest relative to training cost.
Moreover, applying a compression layer-wise helps reduce memory footprint and computational load.
In practice, the compression overhead is not dominant compared to the communication savings achieved in bandwidth-limited environments.

\section{Simulation Results}

\subsection{Models and Dataset}
Our implementation builds upon the Flower framework~\cite{beutel2022flowerfriendlyfederatedlearning}, where we replace the default uniform partitioner with a Dirichlet partitioner~\cite{ribeiro2023federatedlearningcompressiondesigned} to control data heterogeneity via the concentration parameter $\alpha$. Experiments are conducted on ResNet-12, a residual neural network with 780k trainable parameters.
We evaluate on CIFAR-10~\cite{krizhevsky2009learning}, a standard image classification dataset with 50{,}000 32$\times$32 color images across 10 classes, and FEMNIST~\cite{caldas2018leaf}, a federated variant of MNIST, including 814,000 28$\times$28 monochrome images. We consider both IID and non-IID data distributions, where the Dirichlet concentration parameter is set to $\alpha=100$ for the IID case and $\alpha=1$ for the non-IID case. In each communication round, 40\% of the total clients (4 out of 10) are selected for the IID setting, and 20\% (20 out of 100) for the non-IID setting.

\subsection{Hyper-Parameter Tuning}
For CQ, centroid initialization strongly impacts clustering quality and thus overall model performance; among the evaluated methods—Forgy (random), density-based, and linear—linear initialization consistently achieved superior accuracy. We further accelerated the quantization process by adopting CuML’s GPU-optimized $k$-Means implementation~\cite{cuMLKMeans2025}. The best-performing federated configuration used a learning rate of 0.01, a batch size of 8, and one local training epoch per round.

\subsection{Accuracy Analysis} As shown in Table~\ref{tab:accuracies}, IID CIFAR-10 maintains accuracy up to $\gamma=0.5$ and $k=32$, while non-IID CIFAR-10 degrades more sharply, reflecting higher sensitivity to compression. IID FEMNIST remains stable across all compression levels, and non-IID FEMNIST retains accuracy above 0.8 even under aggressive pruning and quantization.
This dataset-dependent behavior arises because, in IID settings (CIFAR-10, FEMNIST), client updates are similar ($\bar{\epsilon}^{(i)} \approx 0$), minimizing compression effects. In non-IID CIFAR-10, diverse updates make $\bar{\epsilon}^{(i)}$ significant, amplifying accuracy loss. In contrast, non-IID FEMNIST benefits from redundancy and partial alignment among updates, keeping $\bar{\epsilon}^{(i)}$ small and preserving accuracy.

\begin{table*}[!ht] 
   \centering 
   \begin{tabular}{c|cccccccccc}
        \toprule
         & \multicolumn{10}{c}{Pruning Rate $\gamma$} \\
        \cmidrule(lr){2-11}
        Number of & 0.1 & 0.2 & 0.3 & 0.4 & 0.5 & 0.6 & 0.7 & 0.8 & 0.9 & 0.95 \\
        \cmidrule(lr){2-11}
        centroids $k$ & \multicolumn{10}{c}{IID / non-IID CIFAR-10,  baseline = 80.7\% / 75.6\%} \\
        \midrule
        256 & 80.6 / 73.6 & 80.3 / 75.0 & 79.8 / 74.5 & 80.0 / 75.0 & 80.1 / 72.0 & 79.9 / 70.6 & 78.6 / 67.5 & 77.9 / 65.5 & 73.9 / 59.3 & 70.3 / 54.3 \\ 
        128 & 81.1 / 73.2 & 80.3 / 73.4 & 80.3 / 72.3 & 80.0 / 70.6 & 79.0 / 69.0 & 79.9 / 66.6 & 79.8 / 63.7 & 77.9 / 59.1 & 73.9 / 55.7 & 70.9 / 50.2 \\ 
        64  & 80.6 / 59.4 & 79.8 / 59.8 & 80.3 / 57.0 & 79.9 / 53.0 & 80.2 / 57.2 & 79.5 / 53.4 & 79.1 / 51.4 & 77.2 / 50.5 & 72.8 / 44.1 & 69.5 / 43.5 \\ 
        32  & 79.2 / 45.4 & 79.2 / 45.8 & 79.1 / 44.9 & 78.9 / 43.6 & 78.5 / 42.5 & 75.5 / 43.8 & 75.7 / 42.1 & 74.8 / 40.3 & 69.6 / 36.2 & 66.9 / 35.5 \\ 
        16  & 70.8 / 38.4 & 69.7 / 39.0 & 70.2 / 39.5 & 70.3 / 38.1 & 69.5 / 36.5 & 68.3 / 37.8 & 67.8 / 36.7 & 66.5 / 34.3 & 64.3 / 34.0 & 61.6 / 31.8 \\ 
        8   & 61.4 / 35.2 & 62.9 / 36.0 & 61.5 / 34.4 & 62.9 / 34.5 & 64.1 / 34.2 & 60.3 / 32.6 & 60.7 / 31.9 & 60.9 / 29.8 & 57.7 / 28.1 & 55.2 / 25.6 \\ 
        4   & 54.6 / 29.4 & 54.1 / 29.0 & 56.7 / 27.3 & 55.0 / 26.4 & 53.8 / 27.9 & 55.0 / 26.9 & 52.9 / 24.7 & 52.1 / 23.4 & 51.4 / 20.4 & 47.3 / 20.0 \\ 
        \midrule
         & \multicolumn{10}{c}{IID / non-IID FEMNIST,  baseline = 88.8\% / 85.7\%} \\
        \midrule
        256 & 88.5 / 85.5 & 88.8 / 85.8 & 88.5 / 86.0 & 88.7 / 85.4 & 88.6 / 85.6 & 88.4 / 85.3 & 88.5 / 85.0 & 88.4 / 84.8 & 88.1 / 83.8 & 88.0 / 82.6 \\ 
        128 & 88.9 / 85.8 & 88.9 / 85.6 & 88.7 / 85.6 & 88.7 / 85.6 & 88.7 / 85.8 & 88.7 / 84.9 & 88.4 / 85.0 & 88.5 / 85.1 & 88.5 / 83.8 & 88.1 / 82.8 \\ 
        64  & 88.7 / 86.0 & 88.7 / 85.9 & 88.9 / 85.7 & 88.6 / 85.8 & 88.6 / 85.4 & 88.7 / 85.3 & 88.6 / 85.0 & 88.4 / 84.5 & 88.2 / 83.6 & 87.8 / 82.7 \\ 
        32  & 88.7 / 85.5 & 88.6 / 85.5 & 88.6 / 84.8 & 88.6 / 84.6 & 88.7 / 85.1 & 88.4 / 84.7 & 88.5 / 84.7 & 88.3 / 84.2 & 88.1 / 83.7 & 87.7 / 82.5 \\ 
        16  & 88.4 / 83.8 & 88.3 / 84.2 & 88.5 / 84.8 & 88.4 / 84.0 & 88.4 / 83.8 & 88.4 / 83.7 & 88.4 / 83.7 & 88.0 / 83.2 & 87.9 / 82.3 & 87.4 / 81.4 \\ 
        8   & 88.1 / 83.4 & 88.1 / 82.7 & 88.2 / 83.0 & 88.2 / 82.9 & 88.1 / 82.9 & 88.1 / 82.7 & 87.8 / 82.7 & 87.9 / 82.2 & 87.5 / 81.0 & 87.0 / 79.8 \\ 
        4   & 87.2 / 80.0 & 87.3 / 80.7 & 87.3 / 80.3 & 87.4 / 80.3 & 87.3 / 80.9 & 87.3 / 80.3 & 87.2 / 79.6 & 87.0 / 79.3 & 86.7 / 78.8 & 86.0 / 76.8 \\ 
        \bottomrule 
   \end{tabular} 
   \caption{FCP performance accuracy values for different pruning rates and quantization levels in multiple scenarios.} 
   \label{tab:accuracies} 
\end{table*}

\subsection{Compression Ratio}
Table~\ref{tab:compression ratio} presents the compression ratios $H_{comm}(\gamma, k)$, representing the client communication overhead in the proposed FCP framework compared to the baseline FL. The results illustrate how different pruning rates and quantization levels affect the amount of data transmitted per communication round. As expected, higher pruning rates and lower quantization levels consistently reduce the communication load, demonstrating that FCP can effectively compress model updates while offering flexibility in selecting compression parameters.

\begin{table*}[!ht]
\centering
\begin{tabular}{cc|cccccccccc}
\toprule
 & Pruning Rate $\gamma$ & 0.1 & 0.2 & 0.3 & 0.4 & 0.5 & 0.6 & 0.7 & 0.8 & 0.9 & 0.95 \\
\midrule
 & 256 & 0.227 & 0.205 & 0.183 & 0.160 & 0.140 & 0.117 & 0.090 & 0.065 & 0.037 & 0.022 \\
 & 128 & 0.193 & 0.179 & 0.158 & 0.139 & 0.122 & 0.101 & 0.079 & 0.056 & 0.032 & 0.019 \\
Number of & 64  & 0.165 & 0.151 & 0.134 & 0.117 & 0.104 & 0.085 & 0.068 & 0.048 & 0.027 & 0.016 \\
centroids $k$ & 32  & 0.133 & 0.122 & 0.111 & 0.097 & 0.086 & 0.073 & 0.057 & 0.043 & 0.023 & 0.014 \\
 & 16  & 0.103 & 0.097 & 0.085 & 0.079 & 0.073 & 0.063 & 0.049 & 0.035 & 0.021 & 0.013 \\
 & 8   & 0.092 & 0.089 & 0.080 & 0.074 & 0.069 & 0.056 & 0.046 & 0.034 & 0.020 & 0.012 \\
 & 4   & 0.084 & 0.079 & 0.073 & 0.067 & 0.061 & 0.054 & 0.043 & 0.032 & 0.018 & 0.011 \\
\bottomrule
\end{tabular}
\caption{Compression ratio for client communication overhead for different pruning rates and quantization levels.}
\label{tab:compression ratio}
\end{table*}

\subsection{Computation Overhead} 
We analyze the computational overhead in terms of client, server, and total computation cost per round. 
Each client performs local training and compression before sending updates to the server. 
The \textit{client overhead} for compression parameters $\gamma, k$ is measured as a time ratio relative to the baseline training cost (without compression): 
\begin{equation}
H_{\text{client}}(\gamma, k) = 
\frac{t_{\text{train}}
+ t_{\text{compress}}
}{t_{\text{train}}
}
.    
\end{equation}

The server performs client selection, decompression, and aggregation. 
The \textit{server overhead} is measured as a ratio relative to the baseline FL server cost (without decompression):
\begin{equation}
H_{\text{server}}(\gamma, k) = 
\frac{t_{\text{select}} + t_{\text{aggregate}} + t_{\text{decompress}} }{t_{\text{select}}
+t_{\text{aggregate}}
}
.    
\end{equation}

Here, $t_{\text{decompress}}$ is CPU-bound and typically much bigger than $t_{\text{aggregate}}$, thus server-side overhead is dominated by decompression.
The ratio of the total computation cost per round relative to baseline, \textit{overall overhead}, is given by:
\begin{equation}
\begin{aligned}
H_{\text{compute}}(\gamma, k) = \frac{t_{\text{compute}}}{t_{\text{select}} + t_{\text{train}}+ t_{\text{aggregate}}},
\end{aligned}   
\end{equation}
with  $t_{\text{compute}} = t_{\text{select}} + t_{\text{train}} + t_{\text{compress}} + t_{\text{decompress}} + t_{\text{aggregate}}$.

Each $H$ denotes an overhead ratio comparing FCP computation overhead with the baseline FL, with $H > 1$ indicating additional overhead introduced by the FCP pipeline. 

After simulation on CIFAR-10 (IID), we observe that the client overhead remains nearly constant ($\sim$1.28--1.1) across pruning rates and quantization levels, suggesting minimal compression cost. In contrast, server overhead decreases sharply (from $\sim$117 to $\sim$26) mostly as $\gamma$ increases, indicating that higher compression substantially reduces server processing. Consequently, the overall computational overhead declines steadily from approximately 2.4 to 1.36, demonstrating that the proposed method achieves significant efficiency gains with negligible impact on computation, especially for clients.

Although relying on GPU for clients is not realistic, this assumption doesn't affect the validity of the FCP. The reason is that the compression cost is small compared to the training cost, and downgrading the computation resources will affect the training much more than the compression.

\subsection{Convergence Speed}
We observe that FL training accuracy curves look much like first order systems step response behavior. Keeping the comparison up, we introduce a metric $\tau$ that corresponds to the round at which the FL training reaches 63\% of its final accuracy. This metric allows an efficient comparison for convergence speed between trainings that don't reach the same final accuracy.

In the FEMNIST settings, the value of $\tau$ remains unchanged, while with CIFAR-10, compression induces a small overhead ($\sim$10\% to 20\%) in terms of convergence speed.

\subsection{FCP Unified Model Cost}
To evaluate the FCP on the global scale, the ratio of the whole training time with the FCP $T_{\text{FCP}}(\gamma,k)$ is compared to that of the baseline $T_{\text{baseline}}$:
\begin{equation}
\rho_{\text{FCP}}(\gamma, k) = \frac{T_{\text{FCP}}(\gamma, k)}{T_{\text{baseline}}}\\ = \frac{5\tau_{\gamma,k}\cdot(t_{\text{compute}}^{\gamma,k}+t_{\text{comm\_down}}+t_{\text{comm\_up}}^{\gamma,k})}{5\tau_{\text{baseline}}\cdot(\frac{t_{\text{compute}}^{\gamma,k}}{H_{\text{compute}}(\gamma,k)}+t_{\text{comm\_down}}+\frac{t_{\text{comm\_up}}^{\gamma,k}}{H_{\text{comm}}(\gamma,k)})}.
\end{equation}

The numerator is the total FCP training time, calculated as $5\tau$ times the per-round execution time. The denominator is computed similarly for the baseline. Downstream communication time is unchanged since FCP only compresses upstream transmissions.

For a practical example, we can use the results of the first experiment (CIFAR-10 IID with 10 clients and 40\% selection rate) with pruning rate $\gamma=0.5$ and $k=32$ quantization clusters. In this setup, the final accuracy drop is around 2\% compared to the baseline. We consider two situations for the network bandwidth available:
\textbf{a.} using a modern Bluetooth technology on low energy mode with $2$ Mbit/s bidirectional rate;
\textbf{b.} using an early 4g technology (LTE Cat 3) with $100$ Mbps download rate and $50$ Mbps upload rate.
We also measure $t_{\text{compute}}^{\text{CIFAR-10 (IID)},0.5,32} = 2.8466\text{s}$ and a model size of 3177~kB before compression (using \texttt{pytorch.save()}) and 274~kB after compression (using the FCP). The convergence speed is $\tau = 14$ compared to $12$ for the baseline.
After calculation, for both situations we find respectively $\rho_{\text{FCP}}(0.5,32)=0.3569$ and $0.9790$, so a global training time reduction of 64\% and 2.1\%. This illustrates the efficiency of the FCP under communication-restrained environments.

\section{Conclusion}

This study tackles the challenges of communication overhead and energy efficiency in FL by introducing the Full Compression Pipeline (FCP), a combination of pruning, quantization, and Huffman encoding. Each component is fine-tuned through detailed analysis and hyperparameter optimization to maximize efficiency with minimal accuracy loss.
Evaluations on models like ResNet-12 trained on CIFAR-10 show that under some conditions, the FCP can significantly reduce size of the model (e.g. by more than 90\%) and speeds up the training rounds about 3$\times$ while maintaining near-baseline accuracy.
Future works may extend the FCP on other model architectures and datasets.
\section*{Acknowledgment}

This work was supported by the Federal Ministry of Research, Technology, and Space (BMFTR, Germany) as part of NeuroSys: Efficient AI-methods for neuromorphic computing in practice (Projekt D) - under Grant 03ZU2106DA, and also German Research Foundation (DFG) under the Cluster of Excellence CARE: Climate-Neutral And Resource-Efficient Construction (EXC 3115), project number 533767731.

\bibliographystyle{IEEEbib}
\bibliography{Refs}

\end{document}